\documentclass{article}

\usepackage[preprint]{nips2023}

\usepackage{siunitx}
\usepackage{pifont}
\usepackage[utf8]{inputenc} 
\usepackage[T1]{fontenc}    
\usepackage{hyperref}       
\usepackage{url}            
\usepackage{booktabs}       
\usepackage{amsfonts}       
\usepackage{nicefrac}       
\usepackage{microtype}      
\usepackage{xcolor}         
\usepackage{adjustbox}
\usepackage{ulem}
\usepackage{multirow} 
\usepackage{makecell}
\usepackage{wrapfig}
\usepackage{xspace}
\usepackage{tcolorbox}
\usepackage{color, colortbl}
\definecolor{citecolor}{HTML}{2980b9}
\definecolor{linkcolor}{HTML}{c0392b}
\definecolor{darkorange}{HTML}{FF8C00}
\definecolor{chocolate}{HTML}{D2691E}
\definecolor{darkgreen}{HTML}{006400}
\definecolor{darkblue}{HTML}{00008B}
\definecolor{mediumblue}{HTML}{0000CD}
\definecolor{dodgerblue}{HTML}{1E90FF}
\definecolor{royalblue}{HTML}{4169E1}
\definecolor{shadecolor}{RGB}{237,237,237}
\definecolor{backred}{RGB}{255, 190, 190}
\definecolor{backblue}{RGB}{210, 230, 250}

\definecolor{zrrgreen}{HTML}{008000}
\definecolor{zrrblue}{HTML}{4682B4}
\definecolor{zrrred}{HTML}{B22222}

\makeatletter
  \newcommand\figcaption{\def\@captype{figure}\caption}
  \newcommand\tabcaption{\def\@captype{table}\caption}
\makeatother

\newcommand{\cmark}{\color[HTML]{036400}{\ding{52}}}%
\newcommand{\xmark}{\color[HTML]{CB0000}{\ding{56}} }%

\newcommand{\eg}{\textit{e.g.}\xspace}

\title{Plot2Code: A Comprehensive Benchmark for Evaluating Multi-modal Large Language Models in Code Generation from Scientific Plots}

\author{%
  Chengyue Wu \\
  The University of Hong Kong\\
  \And
  Yixiao Ge \\
  ARC Lab, Tencent PCG \\
  \texttt{yixiaoge@tencent.com} \\
  \And
  Qiushan Guo \\
  The University of Hong Kong\\
  \And
  Jiahao Wang \\
  The University of Hong Kong\\
  \And
  Zhixuan Liang \\
  The University of Hong Kong\\
  \And
  Zeyu Lu \\
  Shanghai Jiao Tong University \\
  \And
  Ying Shan \\
  ARC Lab, Tencent PCG \\
  \And
  Ping Luo \\
  The University of Hong Kong\\
}

\begin{document}

\maketitle

\begin{abstract}

The remarkable progress of Multi-modal Large Language Models (MLLMs) has attracted significant attention due to their superior performance in visual contexts. 
However, their capabilities in turning visual figure to executable code, have not been  evaluated thoroughly. 
To address this, we introduce \textbf{Plot2Code}, a comprehensive visual coding benchmark designed for a fair and in-depth assessment of MLLMs. We carefully collect 132 manually selected high-quality matplotlib plots across six plot types from publicly available matplotlib galleries. 
For each plot, we carefully offer its source code, and an descriptive instruction summarized by GPT-4.
This approach enables Plot2Code to extensively evaluate MLLMs' code capabilities across various input modalities. Furthermore, we propose three automatic evaluation metrics, including code pass rate, text-match ratio, and GPT-4V overall rating, for a fine-grained assessment of the output code and rendered images. Instead of simply judging pass or fail, we employ GPT-4V to make an overall judgement between the generated and reference images, which has been shown to be consistent with human evaluation.
The evaluation results, which include analyses of 14 MLLMs such as the proprietary GPT-4V, Gemini-Pro, and the open-sourced Mini-Gemini, highlight the substantial challenges presented by Plot2Code.
With Plot2Code, we reveal that most existing MLLMs struggle with visual coding for text-dense plots, heavily relying on textual instruction. 
We hope that the evaluation results from Plot2Code on visual coding will guide the future development of MLLMs. 
All data involved with Plot2Code are available at \url{https://huggingface.co/datasets/TencentARC/Plot2Code}.
\end{abstract}
\section{Introduction}

\begin{figure*}[!h]
    \centering
    \includegraphics[width=0.95\linewidth]{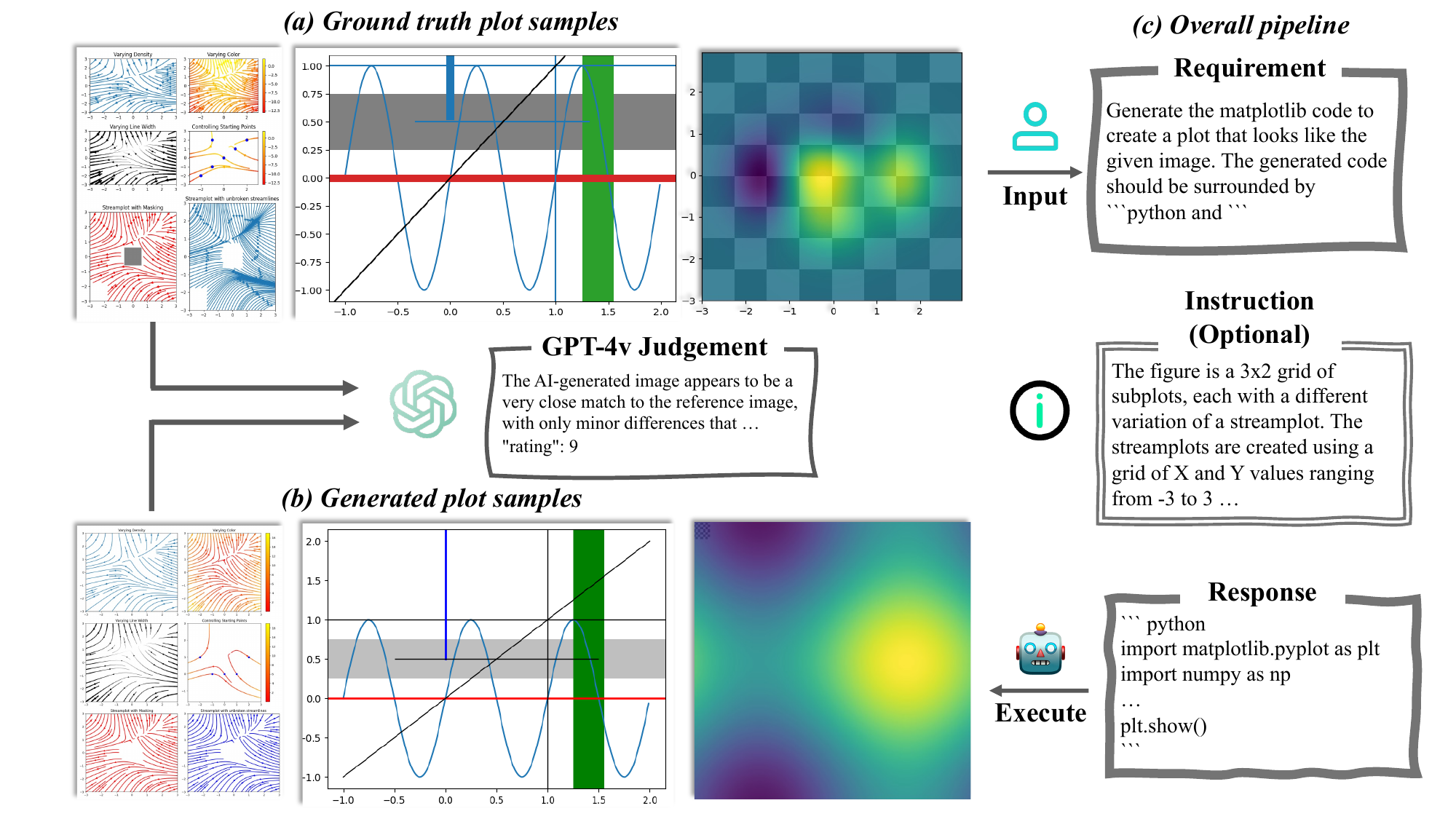}
    \caption{ \textbf{Overview of Plot2Code.}
        \textbf{\textit{Left:}}
        (a) Representative samples from the ground truth plots in our Plot2Code dataset.
        (b) Plot samples generated by multi-modal LLMs using the reference image.
        \textbf{\textit{Right:}}
        (c) The comprehensive pipeline employed to assess the code generation ability of multi-modal LLMs. We consider two distinct settings: \textbf{Direct Asking} and \textbf{Conditional Asking}.
    }
    \label{fig:overall}
\end{figure*}

In the wake of significant advancements in big data and computational power, Large Language Models (LLMs)~\cite{touvron2023llama,brown2020language,hoffmann2022training,kaplan2020scaling}, such as ChatGPT \cite{OpenAI2023ChatGPT} and GPT-4~\cite{OpenAI2023GPT4TR}, have become focal points of interest in both commercial and academic spheres. To extend their versatility across various contexts, Multi-modal Large Language Models (MLLMs)~\cite{ge2024seed,lu2024deepseekvl,openai2023gpt4v} have rapidly evolved, as exemplified by the latest models such as GPT-4V~\cite{openai2023gpt4v}, Gemini~\cite{team2023gemini}, Claude-3~\cite{Claude_3}, and the open-source models LLaVA~\cite{liu2024llavanext,liu2024visual}, Mini-GPT~\cite{zhu2023minigpt,chen2023minigpt} and so on~\cite{ge2024seed,chen2023internvl}. Concurrently, a diverse array of evaluation benchmarks ~\cite{li2023seed,li2023seed2,yue2023mmmu,ying2024mmt} are curated to assess their visual comprehension performance across different domains.
However, there remains a notable gap in attention towards diagrams within text-dense images, which are crucial for assessing the multi-modal reasoning proficiency of MLLMs~\cite{masry2022chartqa,mathew2021docvqa}.

In line with Richard Feynman's philosophy,  "What I cannot create, I do not understand," evaluating the capability of MLLMs to generate code that renders a provided image effectively further showcases their multi-modal understanding and reasoning prowess. 
This particular challenge demands MLLMs to accurately interpret the visual elements present in input diagrams, correlate them with textual context provided, and finally derive executable code to generate the plots. Although the development of code generation from uni-modal natural language has experienced rapid progress in recent years~\cite{roziere2023code,guo2024deepseek,wu2024llama}, the exploration of code generation using multi-modal inputs remains an active area of research. Previous efforts, \eg HumanEval~\cite{chen2021evaluating} and MBPP~\cite{austin2021program}, have concentrated on uni-modal code problems, while the more recent Design2Code~\cite{si2024design2code} has expanded the scope to include user-interface (UI) design, particularly HTML files, for evaluating MLLMs. 
However, these studies focus on unimodal scenarios (\eg, text-only~\cite{chen2021evaluating, austin2021program} or image-only~\cite{si2024design2code} inputs) and have limited capabilities when evaluating models with multimodal inputs.

To this end, our work underscores the critical importance and motivation by addressing three key challenges in evaluating MLLMs' coding capabilities:

\begin{enumerate}
    \item \textbf{Do the evaluation settings accommodate all modalities, including text and images, for both input and output?} This fundamental question pertains to the scope of visual coding. By employing extensive evaluation settings, we can conduct thorough ablation analyses of MLLMs' performance across various input modalities or their combinations, while also assessing the output across different modalities.

    \item \textbf{Are the evaluation metrics accurate, straightforward, and comprehensive?}
    Most existing code benchmarks rely on unit tests to obtain binary evaluation results. While this approach may suffice for uni-modal code tasks, it falls short for visual coding tasks that require not only the code pass rates but also assessments of image fidelity.

    \item \textbf{Are the evaluations for visual coding tasks relevant to real-world applications?} It is imperative that benchmarks align with real-world uses and applications, particularly in coding tasks.
    Employing the commonly used multiple-choice format for evaluating code tasks would be inadequate and incongruous.
\end{enumerate}

Hence, in response to the aforementioned challenges, we present Plot2Code, a comprehensive and specialized multi-modal code benchmark crafted to evaluate the multi-modal understanding, reasoning, and coding capabilities of MLLMs. This benchmark comprises a carefully curated dataset comprising 132 matplotlib plots across 6 plot types, incorporating a total of 293 subplots sourced from matplotlib galleries. Each plot is paired with its corresponding code and a detailed description generated by GPT-4. To cater to diverse input and output formats, Plot2Code includes two evaluation settings, \textbf{Direct Asking} and \textbf{Conditional Asking}, supporting automatic metric-based evaluations for both text and image outputs. MLLMs can be evaluated using text, images, or a blend of both as inputs, while the text and image outputs can be assessed based on the code pass rate and GPT-4V overall rating, which consistently aligns with human evaluations.

We evaluate 14 publicly accessible MLLMs across various evaluation settings to determine optimal performance. Our findings underscore the significant challenges posed by Plot2Code, with GPT-4V only achieving an overall score of 7.68/10, indicating considerable room for enhancement in visual coding tasks.

The contributions of this study can be summarized as follows:
\begin{itemize}
    \item We construct a novel evaluation benchmark, Plot2Code, tailored for multi-modal code tasks, enabling the assessment of advancements in multi-modal understanding and reasoning.
    \item Development of a diverse array of evaluation settings for Plot2Code, accommodating varied modalities for input and output through image-code pairs and automatic evaluation metrics.
    \item Evaluations of various publicly available MLLMs on Plot2Code, revealing that current MLLMs like GPT-4V, Gemini-Pro, and Claude-3, demonstrate modest performance in visual coding tasks.
\end{itemize}

We anticipate that Plot2Code will stimulate the research community to further explore and advance the realm of MLLMs, propelling us towards the realization of truly intelligent multi-modal systems.
\section{Related Work}

\subsection{Multi-modal Large Language Models} With the remarkable progress achieved by Large Language Models (LLMs)~\cite{OpenAI2023ChatGPT,touvron2023llama,OpenAI2023GPT4TR}, incorporating multi-modal input signals into the backbone LLMs has garnered significant interest from both academia and industry~\cite{liu2024llavanext,team2023gemini,lu2024deepseekvl,openai2023gpt4v,team2023gemini}. The primary focus of multi-modal LLM research involves developing additional encoders that enable compatibility and processing of multi-modal inputs by the backbone LLMs. To enhance practical applications, several studies~\cite{wei2023vary, lu2024deepseekvl,team2023gemini} concentrate on processing text-dense images, such as documents and charts, using high-resolution vision encoders. Our goal is to conduct a thorough and comprehensive investigation of MLLMs and their potential by assessing their ability to generate code from reference plots, showcasing their visual coding proficiency.

\subsection{Multi-modal Code Benchmark}

Building upon the growing proficiency of LLMs, a subset of specialized models, known as Code LLMs~\cite{roziere2023code,li2023starcoder,guo2024deepseek}, has emerged. These models focus specifically on programming code, offering numerous appealing applications such as code completion and infilling. Code tasks effectively reflect the in-depth reasoning abilities of (M)LLMs. Uni-modal code benchmarks, like HumanEval and MBPP~\cite{chen2021evaluating,austin2021program}, test the generated code using single-round unit tests with the Pass@k metric. More recently, LLM agents have been evaluated in more complex multi-turn interactive code settings~\cite{wang2023mint,yang2024intercode}. Extending beyond the uni-modal context, MMCode~\cite{li2024mmcode} incorporates image input into code tasks, while Design2Code~\cite{si2024design2code} evaluates MLLMs' generated HTML files through CLIP scores and HTML blocks. Our work proposes a comprehensive benchmark, Plot2Code, which supports a wide range of evaluation scenarios and accommodates both uni-modal and multi-modal inputs. The metrics encompass text-based measures such as code pass rate and generated plot similarity, serving as an all-encompassing evaluation suite for assessing MLLMs' in-depth understanding and reasoning capabilities. See Table \ref{tab:comaprison} for detailed comparisons with related benchmarks.
\section{Dataset Collection}
\begin{figure*}[!h]
    \centering
    \includegraphics[width=\linewidth]{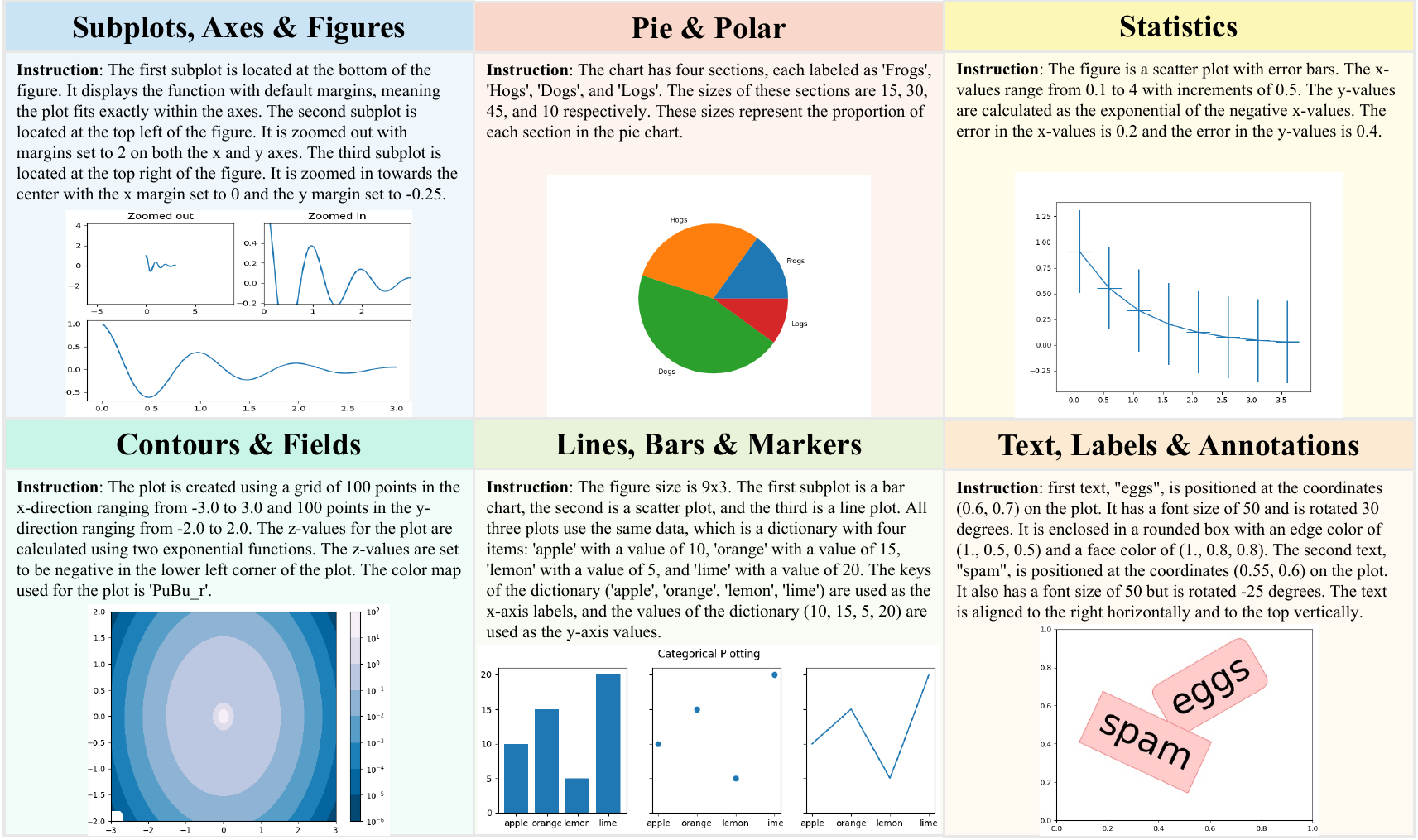}
    \caption{
        \textbf{Examples of Plot2Code benchmark.} We show different-type plots with their corresponding instructions.
    }
    \label{fig:gallery}
\vspace{-5pt}
\end{figure*}
In this section, we delineate the process of curating and processing our benchmark data. Initially, we crawl every website link enumerated in the matplotlib gallery\footnote{\url{https://matplotlib.org/stable/gallery/index.html}}. Subsequently, we extract the code block from each corresponding HTML file. This procedure yields a total of 841 distinct code blocks, which are subjected to further filtering and processing as explicated in the subsequent sections.

\subsection{Test Set Curation}
The primary objective is to acquire well-structured plot-code pairs that effectively evaluate the code generation capabilities of MLLM. It is important to note that the initially crawled Python code may not always be suitable for generating high-quality plots for evaluation purposes. To address this, we employ a combination of automatic processing and manual filtering, as detailed below.

\paragraph{Generation Filtering} During our analysis, we observe that a single HTML file may encompass multiple code segments, with some segments being unable to produce plots due to their focus on import lines and initialization functions. To overcome this limitation, we exclusively extract code from HTML files containing a single code block. This ensures that the extracted code encompasses all essential components and can generate a plot without necessitating additional dependencies. We then filter out all the code that can not generate the plots and obtain 529 plot-code pairs.

\paragraph{Type Filtering} Our analysis operates under the assumption that the plots are simple, static figures devoid of animation and interaction. Consequently, each plot can be regarded as an image file rendered by the matplotlib engine. To maintain this simplicity, we filter out any plots associated with specific tags, such as animation, widget, and event handling, found in their corresponding URLs. A detailed breakdown of the plot-code pair types in our dataset is provided in Figure \ref{fig:type_distribution}.

\paragraph{Manual Curation} Subsequent to the aforementioned processing, we conduct a final round of manual curation to filter examples based on the criteria outlined below:
\begin{itemize} 
\item The plot is devoid of any external file dependencies and can be directly rendered using the corresponding code. 
\item The plots exhibit a wide array of diversity in terms of size, text, colors, and types, thereby serving as a comprehensive benchmark for evaluation that encompasses a wide array of commonly used charts and plots.
\item The plots are uniformly distributed across various difficulty levels, ranging from beginner to specialized levels. 
\end{itemize} 
During the manual filtering process, we adopt a more stringent approach to retain only high-quality plots. Ultimately, we procure 132 test examples that serve as our benchmark.

\subsection{Evaluation Setting}

We assess the test set, curated in the previous step, under two distinct evaluation scenarios: direct asking and conditional asking. To facilitate convenient extraction of code from the MLLM-generated responses, we request the code to be enclosed between specific markers, enabling the use of regular expressions for extraction.

\paragraph{Direct Asking} This setting means giving the MLLM an image as input and requiring it to generate executable code that produces a graph closely resembling the input image. The specific prompt can be found in Appendix ~\ref{sec:code_generation_prompt}. Figure~\ref{fig:direct_asking_case} illustrates an example in this case.

\paragraph{Conditional Asking} 
For MLLMs, this setup means receiving an image and conditions (text instructions) as input and generating executable code that produces results in line with the specified conditions. For LLMs, the input includes conditions only, with other requirements being consistent with those for MLLMs. We employ GPT-4 to extract these instructions from the ground truth code, instructing it to retain all essential information for reproduction while avoiding exposure of code implementation details. The prompt used to construct these instructions can be found in Appendix ~\ref{sec:instruction_generation_prompt}. Figure~\ref{fig:conditional_asking_case} illustrates an example in this case.

\label{sec:setting}
\subsection{Data Statistics}

\begin{figure*}[t]
\centering
\begin{minipage}[c]{0.42\textwidth}
\small
\centering

  \tabcaption{\textbf{Key Statistics of Plot2Code.} Tokens are counted by LLaMA-2 tokenizer.}
  \vspace{2pt}
  \label{tab:statistics}
  \centering
  \begin{adjustbox}{width=\linewidth}
    \begin{tabular}{lr}
    \toprule
    \textbf{Statistic} & \textbf{Number} \\
    \midrule
      Total Samples & 132 \\
      ~- Contours \& Fields & 30 (22.7\%) \\
      ~- Lines, Bars \& Markers & 37 (28.0\%) \\
      ~- Texts, Labels \& Annotations & 14 (10.6\%) \\
      ~- Statistics & 17 (12.9\%) \\
      ~- Subplots, Axes \& Figures & 25 (18.9\%) \\
      ~- Pie \& Polar & 9 (6.8\%) \\
      Total Subplot Count & 293 \\
    \midrule 
      Code Length (tokens) & $401 \pm 281$ \\
      ~- Minimum Length & 60 \\
      ~- Maximum Length & 1823 \\
    \midrule 
      Instruction Length (tokens) & $279 \pm 115$ \\
      ~- Minimum Length & 72 \\
      ~- Maximum Length & 628 \\
    \midrule 
      Text Count & $23 \pm 13 $\\
     \bottomrule
     \end{tabular}
 \end{adjustbox}
\end{minipage}
\qquad
\begin{minipage}[c]{0.5\textwidth}
\centering
\caption{\textbf{Type Distribution of Plot2Code.}}
\label{fig:type_distribution}
\includegraphics[width=\linewidth]{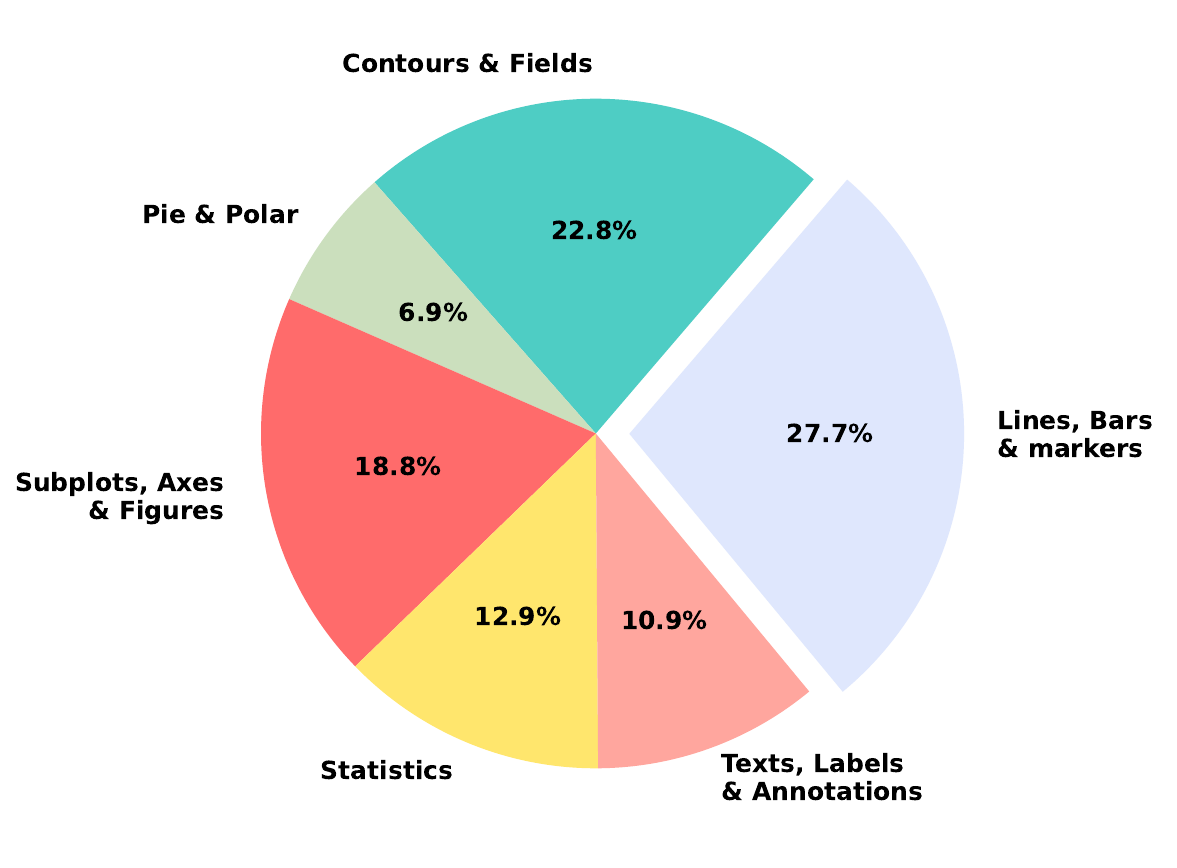}
\end{minipage}
\end{figure*}

\paragraph{Key Statistics} To gauge the difficulty levels of the test examples, we present several statistics in Table \ref{tab:statistics}. We enumerate the total number of subplots present in our test samples, as a single plot may comprise multiple subplots. In total, there are 293 subplots (min=1, max=18). We tokenize the scraped code files using the LLaMA-2 tokenizer~\cite{touvron2023llama}. The ground truth code exhibits an average token count of 409, with a standard deviation of 291. Additionally, we tokenize the instructions accompanying our test samples, yielding an average of 242 tokens and a standard deviation of 58. Utilizing PaddleOCR\footnote{\url{https://github.com/PaddlePaddle/PaddleOCR}}, we calculate the text count for each plot, with an average of 23 and a standard deviation of 13. Collectively, these metrics indicate that our benchmark examples are indeed challenging and encompass a broad spectrum of complexities found in scientific plots.

\paragraph{Type Distribution} To gain insight into the variety of plot types encompassed by our test set, we depict the type distribution via a pie chart in Figure \ref{fig:type_distribution}. For each sample, we determine its type based on the tag present in its URL, sourced from the matplotlib gallery. These categories are defined by the matplotlib gallery itself. The most prevalent types include lines, bars, and markers, while other categories comprise contours, fields, pie charts, polar plots, subplots axes, statistical representations, and text labels and annotations.

\subsection{Evaluation Metrics}
\label{sec:metric}
Unlike unimodal code generation tasks, which can be assessed simply by conducting unit tests and recording the code pass rate, a piece of code might execute flawlessly without any errors but fail to render an image similar to the provided reference image. This discrepancy necessitates the development of precise and automatic evaluation metrics for multimodal code generation tasks. Consequently, we propose a suite of evaluation metrics encompassing various aspects to conduct this evaluation. Our evaluation metrics include the code pass rate, low-level metrics, text match rate, and GPT-4v judgement score. We will elaborate on these metrics in detail in the subsequent paragraphs.

\paragraph{Code Pass Rate} The MLLM is expected to generate code that can be rendered into an image using matplotlib. Consequently, we compute the code pass rate to determine whether the MLLM can generate executable code given the input reference image and instruction. The subsequent metrics will only be applied to images that have been correctly generated.

\paragraph{GPT-4V Judgement} We devise an evaluation pipeline leveraging the GPT-4v model to assess the high-level similarity between generated plots and ground truth plots. This pipeline assigns a rating to test samples on a scale of 1-10, taking into account various aspects such as overall appearance, colors, shapes, positions, and other visual elements of the images. The comprehensive prompt utilized for evaluation can be found in the Appendix ~\ref{sec:eval_prompt}.


\paragraph{Text-Match Ratio} While the high-level similarity evaluation provided by the GPT-4v judgement is valuable, it does not account for detailed plot components such as text, which is crucial for plot interpretation. To compensate for this, we introduce the text-match ratio, aiming to assess the fine-grained similarity between the generated and ground truth plots. This metric evaluates the precision of the text present in the ground truth sample, ensuring that all text elements are accurately reproduced in the plot under assessment and there are no extra texts in the generated image.

\subsection{Comparison with Other Datasets}

As depicted in Table \ref{tab:comaprison}, our dataset encompasses the most extensive range of evaluation settings and metrics compared to all other uni-modal and multi-modal code benchmarks.

\begin{table*}[thb]
\centering

\centering
{\small
\tabcaption{\textbf{Comparison with other uni-modal and multi-modal code benchmarks.}``I" represents images,``T" represents text, and ``I+T" stands for the multi-modal information with images and text.}

\begin{adjustbox}{width=\linewidth}
\begin{tabular}{llccccccccc}
\toprule
\multirow{2}{*}{Dataset} & \multirow{2}{*}{Task Type} & \multicolumn{2}{c}{Input Format} & \multicolumn{3}{c}{Output Eval Format } & \multicolumn{3}{c}{Evaluation Metric} \\ 

& & T & I+T & I & T & I+T & Pass Rate & Component Match & Rating  \\ 
\midrule    

HumanEval~\cite{chen2021evaluating} & Programming & \cmark & \xmark& \xmark& \cmark & \xmark& \cmark & \xmark & \xmark \\

SVGEditBench~\cite{nishina2024svgeditbench} & SVG & \cmark & \xmark& \cmark & \xmark& \xmark& \xmark& \xmark & \xmark &  \\

MMcode~\cite{li2024mmcode} & Algorithm & \cmark & \cmark & \xmark& \cmark & \xmark& \cmark & \xmark     & \xmark \\

Design2Code~\cite{si2024design2code} & Websites & \xmark& \cmark & \cmark & \xmark& \xmark& \xmark& \cmark  & \cmark &  \\

\midrule

Plot2Code & Plots & \cmark & \cmark  & \cmark & \cmark & \cmark & \cmark & \cmark & \cmark \\

\bottomrule
\label{tab:comaprison}

\end{tabular}
\end{adjustbox}
}

\end{table*}
    
\section{Experiments}

\begin{table*}[!t]
\small
\centering
\caption{\textbf{Quantitative results for 14 MLLMs across two settings, Direct Asking and Conditional Asking.} The maximum value of GPT-4V overall rating is \textbf{bolded}.}
\begin{adjustbox}{width=\linewidth}
    \begin{tabular}{l|l|ccc|ccc|}
    \toprule
    \multirow{2}*{\makecell*[l]{Model}}  & \multirow{2}*{\makecell*[l]{Backbone LLM}}   & \multicolumn{3}{c|}{\makecell*[c]{Direct Asking}}
    &\multicolumn{3}{c|}{\makecell*[c]{Conditional Asking}} \\
    & & Pass Rate  & Text-Match  & Rating & Pass Rate  & Text-Match & Rating  \\
    \midrule
    
    \multicolumn{8}{c}{\textit{LLMs}}\\
    \cmidrule{1-8}
    ChatGPT~\cite{OpenAI2023ChatGPT} & ChatGPT~\cite{OpenAI2023ChatGPT} & - & - & - & 80.3 & 56.7 & 6.59 \\
    GPT-4~\cite{OpenAI2023GPT4TR} & GPT-4~\cite{OpenAI2023GPT4TR} & - & - & - &  80.3 & 68.0 & 7.36\\
    GPT-4 (CoT) ~\cite{OpenAI2023GPT4TR} & GPT-4~\cite{OpenAI2023GPT4TR} & - & - & - &  78.8 & 66.0 & 7.09\\
    GPT-4 (PS+) ~\cite{OpenAI2023GPT4TR} & GPT-4~\cite{OpenAI2023GPT4TR} & - & - & - &  77.3 & 66.8 & 7.26\\
    
    \cmidrule{1-8}
    
    \multicolumn{8}{c}{\textit{Closed-source MLLMs}}\\
    \cmidrule{1-8}
    Claude-3-Opus ~\cite{Claude_3}  & Claude-3 ~\cite{Claude_3} & 84.1 & 57.5 & 4.37 & 78.0 & 69.7 & 7.68 \\
    
    Claude-3-Sonnet ~\cite{Claude_3}  & Claude-3 ~\cite{Claude_3} & 75.8 & 46.7 & 5.38 & 65.9 & 57.0 & 7.20 \\
    Gemini-Pro~\cite{team2023gemini}  & Gemini ~\cite{team2023gemini} & 68.2 & 53.6 & 5.06 & 55.3 & 66.9 & 7.10 \\
    GPT-4V~\cite{openai2023gpt4v} &  GPT-4~\cite{OpenAI2023GPT4TR} & 84.1 & 57.7 & \textbf{6.48} &  81.8 & 70.7 & 7.68\\
    GPT-4V (CoT)~\cite{openai2023gpt4v} &  GPT-4~\cite{OpenAI2023GPT4TR} & 89.4 & 56.3 & 6.30 &  81.8 & 69.7 & 7.75\\
    GPT-4V (PS+) ~\cite{openai2023gpt4v} &  GPT-4~\cite{OpenAI2023GPT4TR} & 86.4 & 55.3 & 6.25 &  85.6 & 71.4 & \textbf{7.83}\\
    
    \cmidrule{1-8}
    \multicolumn{8}{c}{\textit{Open-source MLLMs (Low resolution setting) }}\\
    \cmidrule{1-8}
     Mini-Gemini-2B~\cite{li2024mgm} &  Gemma-2B ~\cite{team2024gemma} & 39.4 & 21.4 & 1.96 & 22.7 & 31.8 & 2.80 \\
     Mini-Gemini-8x7B~\cite{li2024mgm}  & Mixtral-8x7B ~\cite{jiang2024mixtral} & 75.8 & 33.9 & 3.76 & 62.1 & 52.3 & 5.74 \\
     Mini-Gemini-34B~\cite{li2024mgm} &  Yi-34B~\cite{young2024yi} & 67.4 & 30.5 & 2.78 & 50.0 & 51.2 & 4.79 \\
    \cmidrule{1-8}
    \multicolumn{8}{c}{\textit{Open-source MLLMs (High resolution setting)}}\\
    \cmidrule{1-8}
     DeepSeek-VL-7B ~\cite{lu2024deepseekvl}  & DeepSeek-7B ~\cite{bi2024deepseek} & 72.0 & 38.7 & 3.69 & 56.8 & 50.1 & 5.19 \\
     LLaVA-1.6-Mistral-7B~\cite{liu2024llavanext}  & Mistral-7B ~\cite{jiang2023mistral} & 64.4 & 32.6 & 3.06 & 42.4 & 45.1 & 4.48 \\
     LLaVA-1.6-34B~\cite{liu2024llavanext}  & Yi-34B ~\cite{young2024yi}& 72.0 & 34.6 & 3.18 & 53.0 & 50.7 & 5.60 \\
     Mini-Gemini-8x7B-HD~\cite{li2024mgm} & Mixtral-8x7B ~\cite{jiang2024mixtral} & 73.5 & 40.7 & 3.87 & 58.4 & 53.7 & 6.08 \\
     Mini-Gemini-34B-HD~\cite{li2024mgm} &  Yi-34B~\cite{young2024yi} & 55.8 & 34.0 & 3.06 & 43.4 & 46.1 & 5.35 \\
     
    \bottomrule
    \end{tabular}
\end{adjustbox}
\label{tab:main-results}
\end{table*}

In this section, we evaluate a variety of multi-modal large language models and methods on our Plot2Code benchmark to compare their performance. This includes both closed-source commercial API models and state-of-the-art open-source models.

\subsection{Evaluation Details}

\paragraph{Evaluated (M)LLMs} To ensure a comprehensive evaluation, we assess 14 representative closed-source and open-source (M)LLMs that vary in parameters, resolution settings, and backbone LLMs, such as GPT~\cite{OpenAI2023ChatGPT}, DeepSeek~\cite{bi2024deepseek}, Mistral~\cite{jiang2023mistral}, Mixtral~\cite{jiang2024mixtral}, and Yi~\cite{young2024yi}. The quantitative evaluation is provided in Sec. We also explore different prompt strategies, including Chain-of-Thought~\cite{wei2022chain} and Plan-and-Solve~\cite{wang2023plan}. ~\ref{sec:quantitative_results}. We investigate the influence of different designs of MLLMs on the performance of our benchmark in Sec. \ref{sec:different_setting_influence}.

\paragraph{Evaluation Methods} As mentioned in Sec.~\ref{sec:setting}, we employ two distinct evaluation settings: Direct Asking and Conditional Asking. For LLMs lacking vision capabilities, we evaluate them solely in the Conditional Asking setting with instruction input. Furthermore, we extend the GPT-4V judgement setting to conduct pairwise evaluations between two (M)LLMs and perform a correlative analysis between GPT-4V judgement and human evaluation. More details are provided in Sec.~\ref{sec:compare_pairs}.

\subsection{Overall Evaluation}
\label{sec:quantitative_results}

We showcase the quantitative results of (M)LLMs on our Plot2Code benchmark here. The code pass rate, text-match ratio, and GPT-4V overall rating for both direct asking and conditional asking scenarios are reported in Table ~\ref{tab:main-results}.

\paragraph{The comprehensive challenge of Plot2Code.} The benchmark poses considerable challenges, as even advanced models like Claude-3-Opus, Gemini-Pro, and GPT-4V achieve only 7.68, 7.10, and 7.68, respectively, in the overall assessment for the conditional asking scenario, indicating substantial room for improvement. In addition to the overall rating, the pass rate also presents challenges for MLLMs, particularly when instructions are added. For example, Gemini-Pro's pass rate decreases from 68.2\% to 55.3\% after incorporating the instruction, as the added requirements will make it harder to generate the corresponding code. In contrast to widely used benchmarks like MT-bench and HumanEval, where recent advanced models attain ratings above 9.00 and code pass rates exceeding 80\%, Plot2Code necessitates both visual understanding and reasoning abilities to analyze the plot, generate executable code, and create a plot resembling the reference plot. This heightened challenge for (M)LLMs serves as a more rigorous examination of their visual reasoning comprehension and reasoning capabilities.

\paragraph{The gap between closed-source and open-source models.} The performance of open-source models lags considerably behind that of closed-source models. We evaluate recently advanced open-source MLLMs, including DeepSeek-VL~\cite{lu2024deepseekvl}, Mini-Gemini~\cite{team2023gemini}, and LLaVA-Next~\cite{liu2024llavanext}. The best performance among the open-source MLLMs is achieved by Mini-Gemini-8x7B-HD, which scores 6.08 in the GPT-4V judgement with a 58.4\% code pass rate. However, this performance is still not on par with that of commercial closed-source MLLMs. There is a need for the open-source community to develop more powerful models that can compete with, or even surpass, the capabilities of advanced proprietary models.

\subsection{Influence of difference settings. }
\label{sec:different_setting_influence}

 We analyze the results from various perspectives, encompassing prompt strategies, backbone LLMs, and the resolution settings. The key findings are summarized as follows.
 
\paragraph{The influence of LLMs. } As depicted in Table \ref{tab:main-results}, there is a strong correlation between model performance and the backbone LLM used, evident in both Mini-Gemini and LLava. This suggests that the Plot2Code task may require powerful backbone LLMs to facilitate the reasoning process and generate executable code.

\paragraph{Different evaluation settings. } As mentioned in Sec. \ref{sec:setting}, there are two distinct evaluation settings. In Table \ref{tab:main-results}, it can be observed that in the conditional asking setting, MLLMs generally achieve a lower pass rate and higher similarity compared to the direct asking setting. We attribute this to the fact that the added instruction imposes stricter restrictions or requirements on the generated code, making it more challenging for models to generate executable code. However, the additional instruction can enhance the similarity of the generated image to the reference image. Beyond these two evaluation settings, we also investigate the influence of different prompt strategies, including Chain-of-Thought and Plan-and-Solve. We find that prompts encouraging MLLMs to engage in deeper thinking do not show a clear advantage over our default prompt, indicating that the exploration of multi-modal reasoning prompts is still in progress.

\begin{figure*}[t]
\centering
\begin{minipage}[c]{0.45\textwidth}
\centering
\caption{\textbf{Pair evaluation results in the conditional asking setting.}~We use GPT-4V without prompt strategies as the baseline (this method is not shown in the table as it serves as the basis for pairwise comparison).}
\label{fig:pair_evaluation}
\vspace{5pt}
\includegraphics[width=\linewidth]{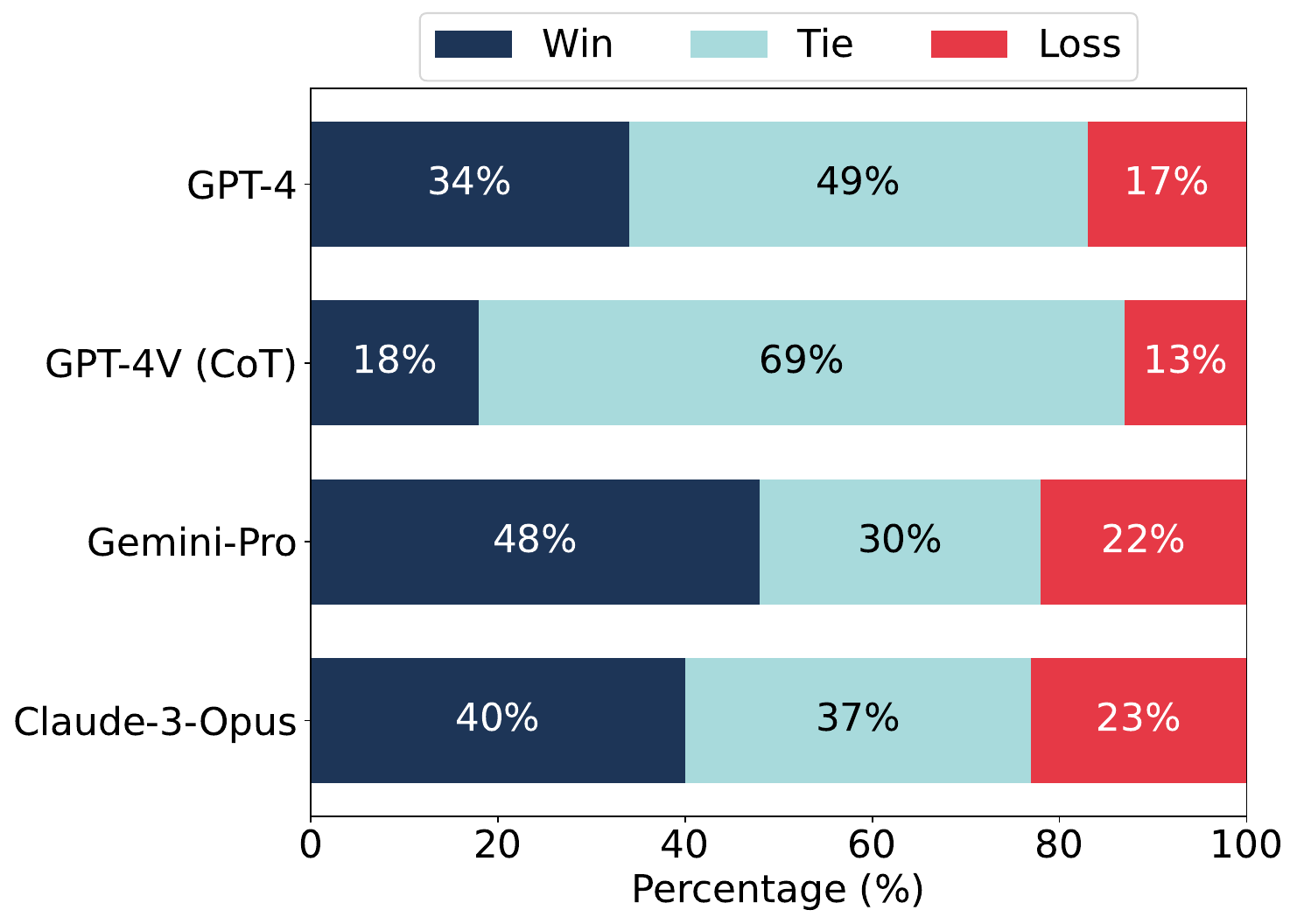}
\end{minipage}
\qquad
\begin{minipage}[c]{0.45\textwidth}
\centering
\caption{\textbf{Ablation experiments involving the addition of OCR tokens or not.}~The base model is Mini-Gemini-8x7B-HD. OCR tokens are extracted using PaddleOCR, which is supported by the official implementation of Mini-Gemini.}
\label{fig:ocr_experiment}
\vspace{5pt}
\includegraphics[width=\linewidth]{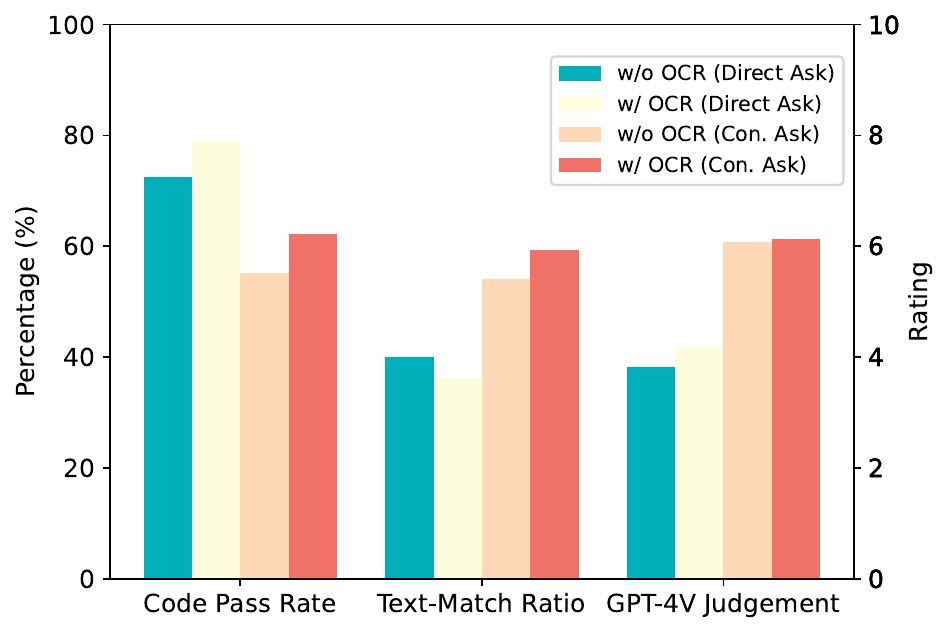}
\end{minipage}
\vspace{-5pt}
\end{figure*}

\paragraph{Different image resolution settings.} In addition to the backbone LLMs, we examine the vision encoder settings, specifically focusing on image resolution settings. Many MLLMs utilize vision encoders with higher resolution capabilities to provide more detailed information from the input images. We observe that MLLMs with higher resolution consistently improve performance, a trend similar to that seen in ChartQA \cite{masry2022chartqa} and DocQA~\cite{mathew2021docvqa}. We also conduct an experiment wherein we add OCR tokens extracted by the official PaddleOCR implementation used in Mini-Gemini, as shown in Figure \ref{fig:ocr_experiment}. This approach yields performance improvements akin to those achieved with high-resolution settings, suggesting current MLLMs may still require more powerful vision encoders to capture detailed information from images.

\subsection{Pairwise Model Comparison}

\label{sec:compare_pairs}
In accordance with the conventional practice of pairwise model evaluation~\cite{zheng2024judging, zhou2024lima}, we expand the GPT-4V judgement setting to perform pairwise evaluations between two (M)LLMs. The detailed prompt employed for pairwise model comparison can be found in the Appendix ~\ref{sec:eval_prompt}. For each reference sample, we request GPT-4V to determine which generated image is more similar when comparing a pair of MLLMs. To mitigate the influence of differing positions, we swap the two generated images for an additional evaluation. A model is considered victorious only if it wins both rounds; otherwise, the result is deemed a tie. We utilize GPT-4V as the baseline for comparison. The results are illustrated in Figure \ref{fig:pair_evaluation}, from which we can infer that: (i) Compared to GPT-4, the inclusion of image input for GPT-4V is beneficial in generating higher quality plots. (ii) The commonly used prompt strategy, Chain-of-Thought, does not yield additional advantages in our benchmark.
\section{Statistical Analysis}

\begin{figure*}[t]
\centering
\begin{minipage}[c]{0.47\textwidth}
\small
\centering

  \tabcaption{\textbf{Comparison of different indicators in distinguishing image groups with and without significant differences.}}
  \vspace{3pt}
  \centering
  \begin{adjustbox}{width=\linewidth}
    
    \centering
\begin{tabular}{@{}lccS@{}} 
\toprule
\textbf{Indicators}       & \textbf{t-statistic} & \textbf{p-value} & \textbf{Reject $H_0$} \\ \midrule
MSE                       & -1.24                & \num{0.22}       & \xmark             \\
SSIM                      & 1.28                 & \num{0.21}       & \xmark             \\
CLIP-Score                & 4.23                 & \num{1.24e-4}    & \cmark             \\
Text-Match Ratio          & 5.69                 & \num{9.62e-7}    & \cmark             \\
GPT-4V Judgement          & \textbf{9.07}        & \num{1.22e-11}   & \cmark    \\ \bottomrule
\end{tabular}
    
 \end{adjustbox}
\end{minipage}
\qquad
\begin{minipage}[c]{0.4\textwidth}
\centering
\caption{\textbf{Illustration of inaccuracy of traditional low-level metrics.}}
\vspace{5pt}
\label{fig:hypothesis_test}
\includegraphics[width=\linewidth]{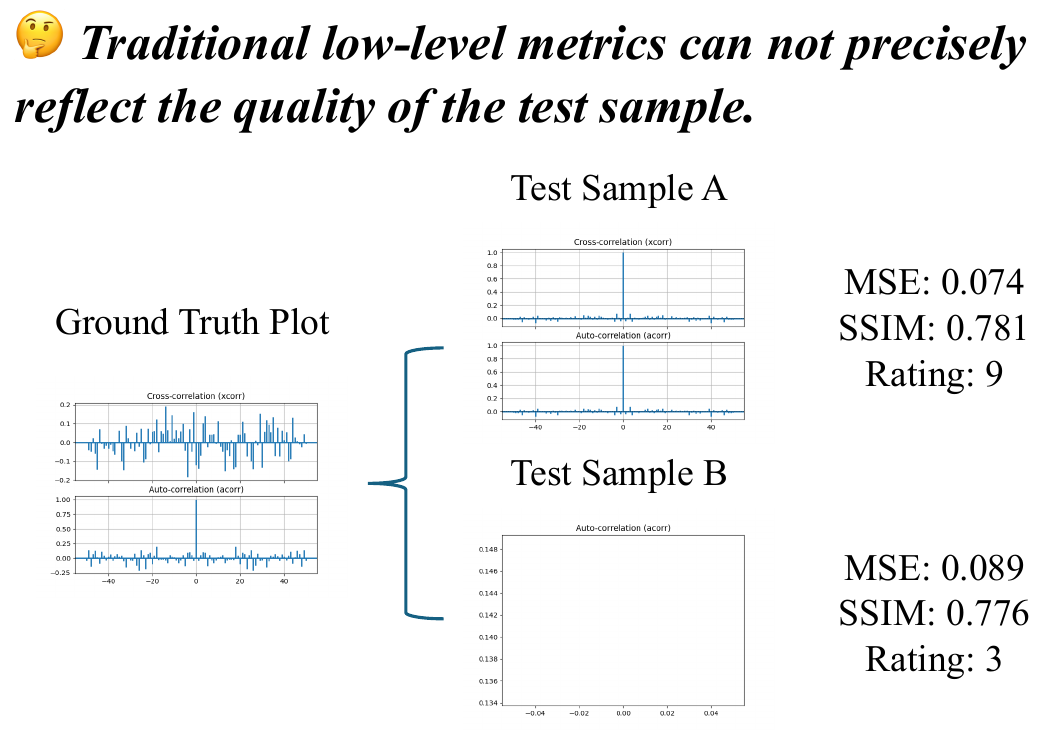}
\end{minipage}
\end{figure*}
In this section, we perform several statistical analyses to justify the design of our benchmark. Since our benchmark involves comparing the generated image with the reference image, we first investigate whether the proposed metrics can effectively indicate image similarity. Additionally, we conduct a correlative analysis to determine the relationship between GPT-4v judgement and human evaluation, further substantiating the basic intuition behind our proposed metric and benchmark.

\paragraph{Hypothesis tests for image similarity metrics.} Since plot images typically exhibit similarity due to their common white background and overall structure, hypothesis tests are conducted to evaluate the effectiveness of different indicators in distinguishing between image groups with and without visual differences. We first select two image groups: one group consists of images generated by GPT-4V with high visual similarity, while the other group contains images from Mini-Gemini-2B with considerably lower similarity. We then perform a hypothesis test using several metrics, including traditionally used Mean Squared Loss (MSE), Structural Similarity (SSIM), as well as the metrics we proposed, namely text-match ratio and GPT-4V rating, to determine whether the metric can effectively differentiate between the two image groups with varying levels of visual similarity.

Specifically, we conduct two-sample t-test (independent t-test). 

\begin{equation}
t = \frac{\bar{X}_1 - \bar{X}_2}{\sqrt{\frac{s_1^2}{n_1} + \frac{s_2^2}{n_2}}}
\end{equation}

where $\bar{X}_1$ and $\bar{X}_2$ are the sample means of the two groups, $s_1^2$ and $s_2^2$ are the sample variances, and $n_1$ and $n_2$ are the sample sizes. The null hypothesis ($H_0$) assumed that there was no significant difference in the scores of the indicators between the two image groups. The alternative hypothesis ($H_1$) assumed that there was a significant difference. The p-value, a measure of the strength of evidence in support of a null hypothesis, was calculated for each indicator. A smaller p-value indicates stronger evidence against the null hypothesis. In this case, a significance level of 0.05 was chosen, meaning that if the p-value was less than 0.05, the null hypothesis would be rejected in favor of the alternative hypothesis, with a 95\% probability of correctly rejecting the null hypothesis. The results of our hypothesis test can be found in Table ~\ref{fig:hypothesis_test}.

For the MSE and SSIM indicators, the p-values were 0.22 and 0.21, respectively, both of which were greater than the significance level. Therefore, we could not reject the null hypothesis for these two indicators, suggesting that they might not effectively distinguish between the two image groups.

On the other hand, for the GPT-4V-Judgement and Text-Match Ratio indicators, the p-values were $1.22\times10^{-11}$ and $9.62\times10^{-7}$, respectively, both of which were less than the significance level and CLIP-score's p-value ($1.24\times10^{-4}$). Therefore, we rejected the null hypothesis for these two indicators, suggesting that they could effectively distinguish between the two image groups.

It should be noted that hypothesis testing only provides statistical evidence based on the collected data, and does not prove whether the null hypothesis or the alternative hypothesis is absolutely true.


\begin{table}[t]
\small
\centering
\caption{\textbf{Correlation coefficient comparison between GPT evaluations and human evaluations.}}
\resizebox{0.75\textwidth}{!}{
\begin{tabular}{@{}cccccc@{}}
\toprule
\multicolumn{2}{c}{\textbf{Kendall's Tau}} & \multicolumn{2}{c}{\textbf{Pearson}} & \multicolumn{2}{c}{\textbf{Spearman}} \\ \cmidrule(lr){1-2} \cmidrule(lr){3-4} \cmidrule(lr){5-6}
                      Coefficient       & p-value              & Coefficient         & p-value         & Coefficient           & p-value          \\ \midrule
                   0.437             & $8.68\times10^{-49}$ & 0.479 & $6.89\times10^{-54}$ & 0.469 & $1.57\times10^{-51}$ \\ \bottomrule
\end{tabular}}
\label{tab:correlations}
\vspace{-5pt}
\end{table}
\paragraph{Correlative analysis between GPT-4V judgement and human evaluation.} To investigate the similarity between the GPT-4V judgement and human evaluation, a correlative analysis was performed using different correlation coefficients, including Kendall's Tau, Pearson correlation coefficient, and Spearman's rank correlation coefficient. The sample size for this analysis was 920. Details can be found in the Appendix ~\ref{sec:appendix_correlation_analysis}.

As shown in Table \ref{tab:correlations}, All three correlation coefficients indicated a moderate positive relationship between the GPT-4V judgement and human evaluation. Moreover, the p-values were all smaller than the significance level of 0.05, suggesting that the correlations were statistically significant.

These findings imply that the GPT-4V judgement is in general agreement with human evaluation, demonstrating its effectiveness in assessing the similarity between generated images and real images.

\section{Conclusion}

In this study, we have presented a comprehensive benchmark, Plot2Code, for the evaluation of multi-modal language model's code generation ability. This benchmark encompasses a wide range of complexities and types of scientific plots, making it a robust tool for assessing the performance of different models. We have proposed a suite of evaluation metrics, including code pass rate, text-match ratio, and GPT-4v judgement score, which together provide a holistic evaluation of a model's performance.

Our evaluation of various models on the Plot2Code benchmark has revealed significant differences in performance, highlighting the challenges posed by this task and the room for improvement in current models. We have found that while some models can generate executable code and produce plots similar to the reference image, accurately reproducing all text elements and fine-grained details remains a challenge.

In future work, we believe our Plot2Code benchmark can stimulate further exploration of multi-modal reasoning, text-dense image understanding, and complex code generation capabilities of MLLMs. There are numerous aspects that are worth exploring, including multi-modal reasoning prompt strategies, the design of vision encoders that are compatible with text-dense images. 
In future work, we believe our Plot2Code benchmark can stimulate further exploration of multi-modal reasoning, text-dense image understanding, and complex code generation capabilities of MLLMs. There are numerous aspects that are worth exploring, including multi-modal reasoning prompt strategies, the design of vision encoders that are compatible with text-dense images. We hope more research related to this aspects can further reduce the gap between the open-source community MLLMs and closed-source commercial APIs.

\normalem
{\small
\bibliographystyle{plain}
\bibliography{ref}
}
\newpage
\appendix
\onecolumn
\section{Prompt Template}

In this section, we introduce the prompt template used for the experiment.

\subsection{Prompt for Code Generation}
\label{sec:code_generation_prompt}
We use the following for the direct asking setting.

\begin{tcolorbox}
You are a helpful assistant that can generate Python code using matplotlib. Generate the matplotlib code to create a plot that looks like the given image, as similar as possible. The generated code should be surrounded by \textasciigrave \textasciigrave \textasciigrave python and \textasciigrave \textasciigrave \textasciigrave

\vspace{0.5cm}
\textit{<image\_token>}\textit{<image\_token>}\textit{<image\_token>}...
\end{tcolorbox}

For the conditional asking setting, we add the instruction of the reference plot at the front of the direct asking prompt.

\begin{tcolorbox}
\textit{<instruction>}
\vspace{0.5cm}

You are a helpful assistant that can generate Python code using matplotlib. Generate the matplotlib code to create a plot that looks like the given image, as similar as possible. The generated code should be surrounded by \textasciigrave \textasciigrave \textasciigrave python and \textasciigrave \textasciigrave \textasciigrave

\vspace{0.5cm}
\textit{<image\_token>}\textit{<image\_token>}\textit{<image\_token>}...
\end{tcolorbox}

\subsection{Prompt for Instruction Generation}
\label{sec:instruction_generation_prompt}

Here is the prompt for generating each plot's corresponding instruction. We requite the GPT-4 to examine the code for each plot and summarize the key information in it without any implementation details.

\begin{tcolorbox}
Please review the Python code provided below, which uses matplotlib.pyplot to generate figures. Your job is to identify key details, like type, texts, etc., required to recreate a figure from the given code: \textit{<code>}

\vspace{0.5cm}

Remember, your response should not include any code and avoid implementation details. Do not describe any detailed variables or functions in the code. Instead, use everyday language to describe the necessary information. If the code uses random seed, you should extract it for reproduction. Reveal the data used in the figure for recreation. Strictly follow the rule that do not expose any variables or functions used in the code. Summarize the crucial information as follows:
\end{tcolorbox}

\subsection{Prompt for Evaluation}
\label{sec:eval_prompt}

We use the following prompt for GPT-4V overall rating. We will provide both the ground truth image and the test image generated by the MLLM assistant for GPT-4V to rate the similarity.

\begin{tcolorbox}
You are a helpful assistant. Please evaluate the similarity between a reference image created using matplotlib and an image generated by code provided by an AI assistant. Consider factors such as the overall appearance, colors, shapes, positions, and other visual elements of the images. Begin your evaluation by providing a short explanation. Be as objective as possible. After providing your explanation After providing your explanation, you must rate the response on a scale of 1 to 10 by strictly following this format: "[[rating]]", for example: "Rating: [[5]]",

\vspace{0.5cm}
\textit{<gt\_image\_token>}\textit{<gt\_image\_token>}\textit{<gt\_image\_token>}...

\vspace{0.5cm}
\textit{<test\_image\_token>}\textit{<test\_image\_token>}\textit{<test\_image\_token>}...
\end{tcolorbox}

In the pair evaluation, we utilize the following prompt to determine which generated image, either from Assistant A or Assistant B, is more similar to the ground truth image.

\begin{tcolorbox}
You are a helpful assistant. Please act as an impartial judge and evaluate the quality of the generated images provided by two AI assistants given the ground truth image displayed below. You should choose the assistant that generate the more similar image. Your evaluation should consider factors such as the overall appearance, colors, shapes, positions, and other visual elements of the images.

\vspace{0.5cm}

Here is the ground truth image.

\vspace{0.5cm}
\textit{<gt\_image\_token>}\textit{<gt\_image\_token>}\textit{<gt\_image\_token>}...
\vspace{0.5cm}

Here is the image generated by the assistant A.

\vspace{0.5cm}
\textit{<test\_image\_A\_token>}\textit{<test\_image\_A\_token>}\textit{<test\_image\_A\_token>}...
\vspace{0.5cm}

Here is the image generated by the assistant B.

\vspace{0.5cm}
\textit{<test\_image\_B\_token>}\textit{<test\_image\_B\_token>}\textit{<test\_image\_B\_token>}...
\vspace{0.5cm}

Begin your evaluation by comparing the two responses and provide a short explanation. Avoid any biases and ensure that the order in which the responses were presented does not influence your decision.
\end{tcolorbox}

\subsection{Prompt Strategy}
\label{sec:prompt_strategy}

Additionally, we experiment with alternative prompt strategies that promote more reasoning by MLLM assistants, such as Chain-of-Thought~ (CoT) \cite{wei2022chain} and Plan-and-Solve (PS) ~\cite{wang2023plan}.

The Chain-of-Thought (CoT) prompt is demonstrated below, wherein a specific sentence is added at the commencement of the assistant's response.

\begin{tcolorbox}
\textit{<USER>}

\vspace{0.5cm}
You are a helpful assistant that can generate Python code using matplotlib ...
\tcblower
\textit{<Assistant>}

\vspace{0.5cm}
Let us think step by step. ...
\end{tcolorbox}

We modify Plan-and-Solve (PS) strategy to make it compatible with our visual coding task and call it as PS+ in Table ~\ref{tab:main-results}. It first encourage MLLM assistants to make a detailed plan.

\begin{tcolorbox}
\textit{<USER>}

\vspace{0.5cm}
You are a helpful assistant that can generate Python code using matplotlib ...
\tcblower
\textit{<Assistant>}

\vspace{0.5cm}
Let us first describe the plot and make a detailed plan step by step ...
\end{tcolorbox}

If the assistant outputs the code during the first step, the strategy will terminate. Otherwise, the second step will be employed, prompting the assistant to produce the final answer based on the plan described in the first stage.

\begin{tcolorbox}
\textit{Previous messages...}
\tcblower
\textit{<Assistant>}

\vspace{0.5cm}
Based on the above description, now we are prepared to generate the code. The generated code is surrounded by \textasciigrave \textasciigrave \textasciigrave python and \textasciigrave \textasciigrave \textasciigrave to make it easier to be extracted by regular expressions. Therefore, the code is:

\end{tcolorbox}

\section{Case Study}

\begin{figure*}[!h]
    \vspace{-2cm}
    \centering
    \includegraphics[width=0.9\linewidth]{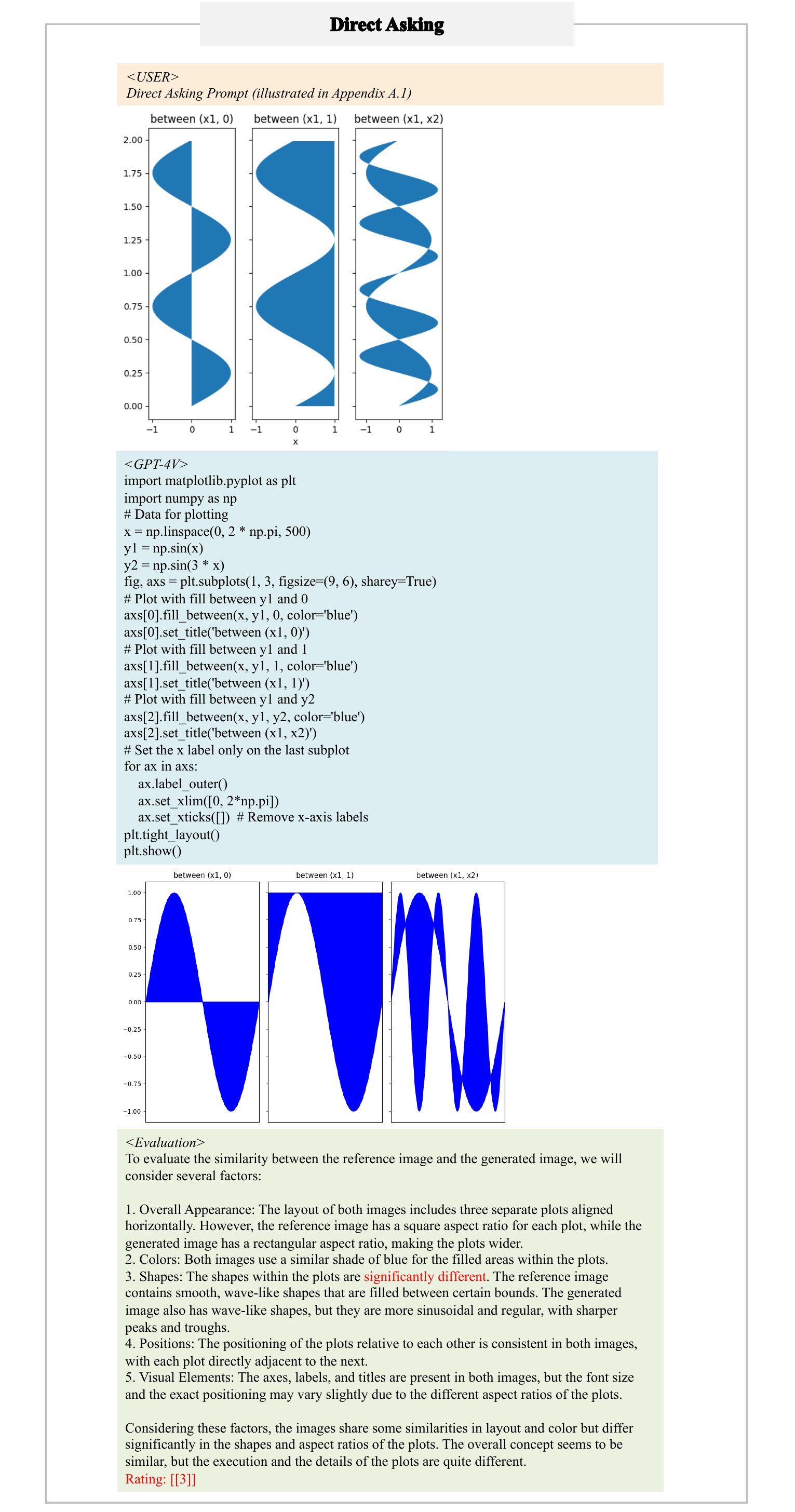}
    \caption{A case of Direct Asking, showcasing the generated code, plot, and evaluation result.}
    \label{fig:direct_asking_case}
\end{figure*}

\begin{figure*}[!h]
    \vspace{-2cm}
    \centering
    \includegraphics[width=0.78\linewidth]{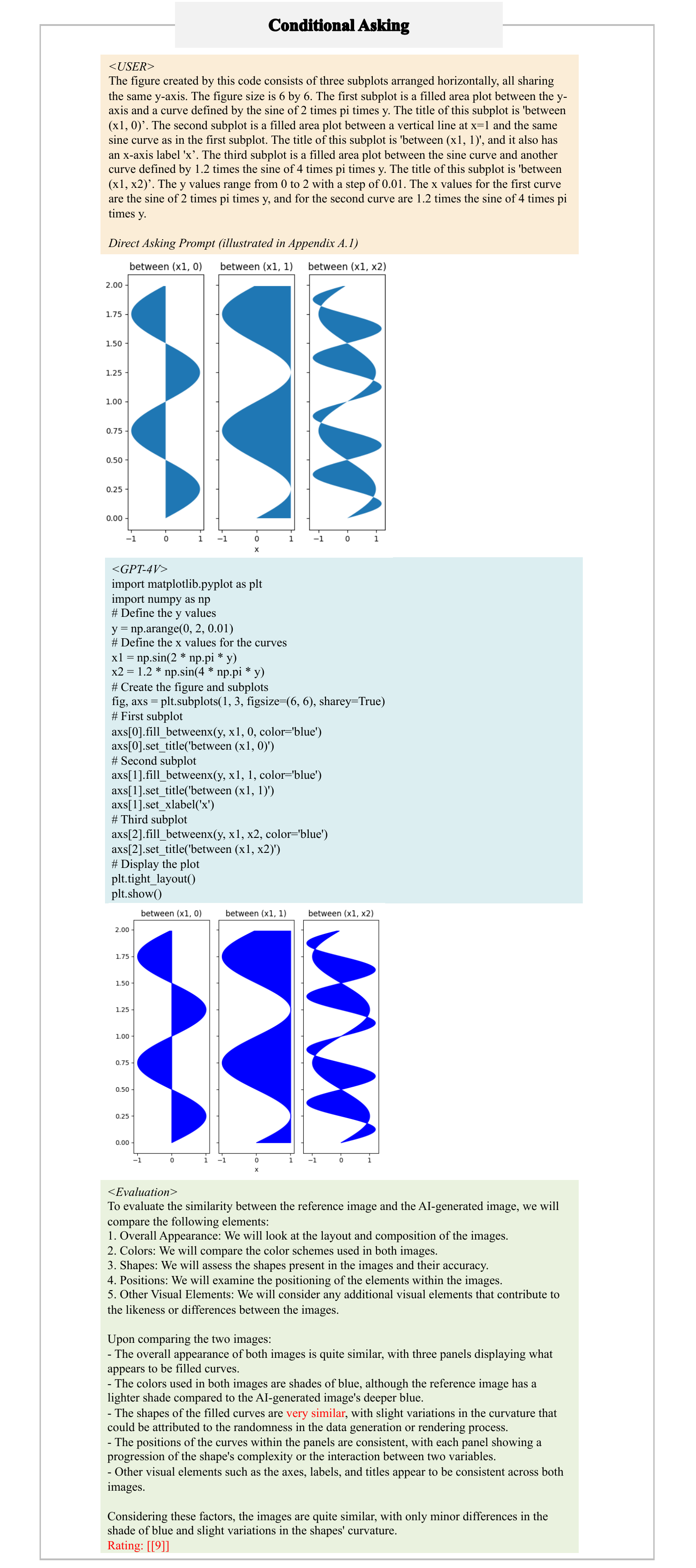}
    \caption{A case of Conditional Asking, showcasing the generated code, plot, and evaluation result.}
    \label{fig:conditional_asking_case}
\end{figure*}

\begin{figure*}[!h]
    \vspace{-2cm}
    \centering
    \includegraphics[width=0.9\linewidth]{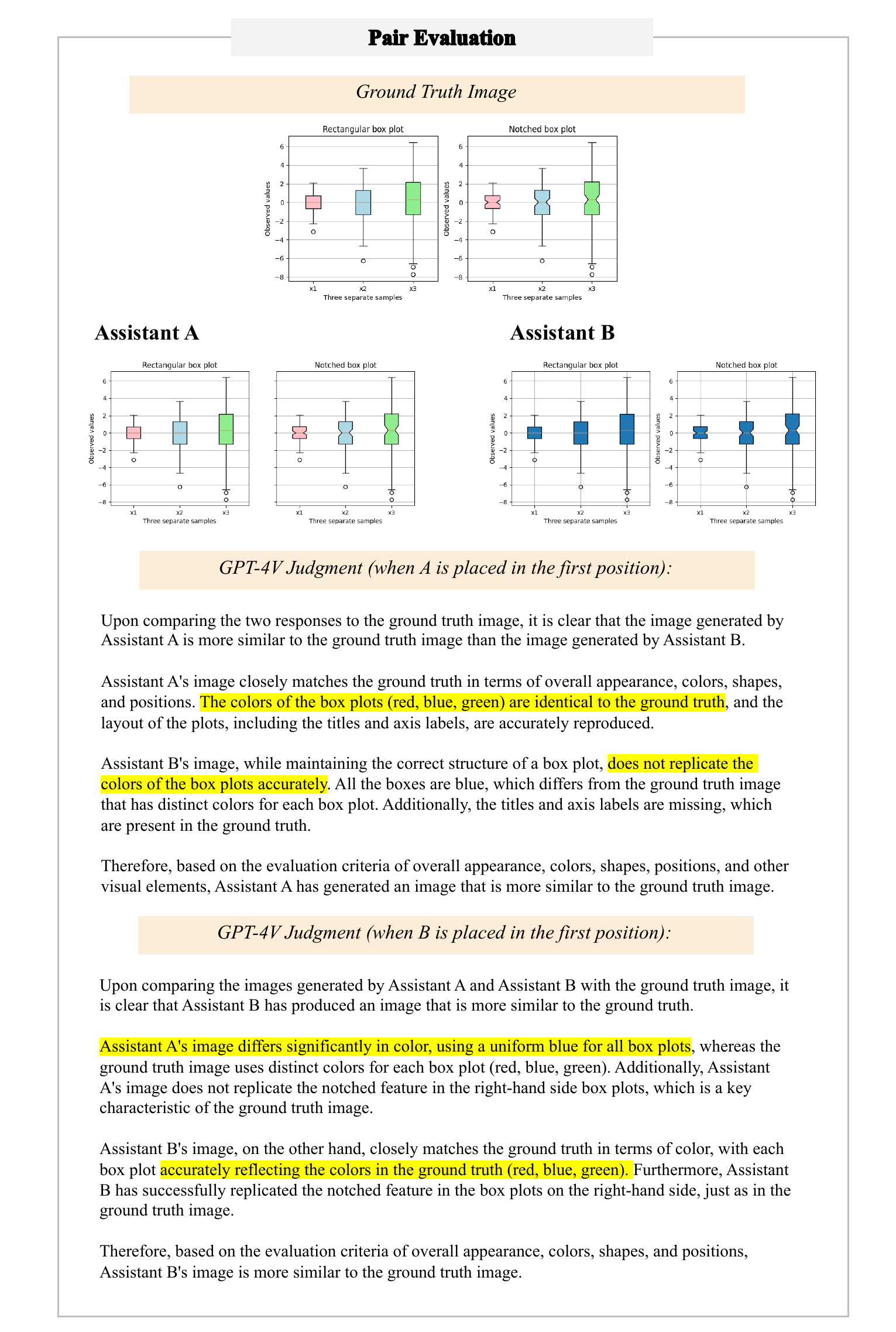}
    \caption{A case of pair evaluation. We interchange the order of responses from the two assistants and conduct the evaluation twice. Both results are presented here.}
    \label{fig:pair_evaluation_case}
\end{figure*}

In this section, we present several examples using GPT-4V as the model under evaluation. We showcase cases from the direct asking setting (Figure \ref{fig:direct_asking_case}), the conditional asking setting (Figure \ref{fig:conditional_asking_case}), and the pair-evaluation setting compared to Gemini-Pro (Figure \ref{fig:pair_evaluation_case}), respectively. All the samples are drawn with the default prompt strategy.

\section{Correlation Analysis}
\label{sec:appendix_correlation_analysis}
In this section, we discuss the details of the correlation analysis. We select 20 pair evaluation samples with GPT-4v as the baseline in the conditional asking setting (10 compared to Gemini-Pro, 10 compared to Claude-3-Opus). Subsequently, we use these 20 samples to create an online questionnaire and invite colleagues from the lab, who hold at least a bachelor's degree, to participate. Each question in the questionnaire presents the ground truth image, the generated image from Assistant A, and the generated image from Assistant B. Participants are asked to choose one of the following three options: 
\begin{itemize}
    \item Assistant A's generated image is more similar to the ground truth image
    \item Assistant B's generated image is more similar to the ground truth image
    \item the level of similarity is close
\end{itemize}
In the end, we receive 46 completed questionnaires, resulting in 46 x 20 = 920 samples for conducting the correlation analysis.


\end{document}